\documentclass[runningheads]{llncs}
\usepackage[T1]{fontenc}
\usepackage{graphicx}
\usepackage{algorithm}
\usepackage{algorithmic}
\usepackage{amsmath}
\usepackage[misc]{ifsym}
\begin{document}
\title{Towards Point Cloud Compression for Machine Perception: A Simple and Strong Baseline by Learning the Octree Depth Level Predictor
}
\titlerunning{PCCMP-Net}
%
\author{Lei Liu\inst{1}{(\textrm{\Letter})} 
\and
Zhihao Hu\inst{1} \and
Zhenghao Chen\inst{2}}
\authorrunning{Lei Liu et al.}
\institute{School of Computer Science and Engineering, Beihang University,  China \and
School of Electrical and Information Engineering, \\The University of Sydney, Australia}
\maketitle 
\renewcommand{\thefootnote}{\fnsymbol{footnote}} 
\footnotetext[0]{\textrm{\Letter} Corresponding author: Lei Liu (liulei95@buaa.edu.cn)} 
\vspace{-10mm}
\begin{abstract}
Point cloud compression has garnered significant interest in computer vision. However, existing algorithms primarily cater to human vision, while most point cloud data is utilized for machine vision tasks. To address this, we propose a point cloud compression framework that simultaneously handles both human and machine vision tasks. Our framework learns a scalable bit-stream, using only subsets for different machine vision tasks to save bit-rate, while employing the entire bit-stream for human vision tasks.
Building on mainstream octree-based frameworks like VoxelContext-Net, OctAttention, and G-PCC, we introduce a new octree depth-level predictor. This predictor adaptively determines the optimal depth level for each octree constructed from a point cloud, controlling the bit-rate for machine vision tasks. For simpler tasks (\textit{e.g.}, classification) or objects/scenarios, we use fewer depth levels with fewer bits, saving bit-rate. Conversely, for more complex tasks (\textit{e.g}., segmentation) or objects/scenarios, we use deeper depth levels with more bits to enhance performance.
Experimental results on various datasets (\textit{e.g}., ModelNet10, ModelNet40, ShapeNet, ScanNet, and KITTI) show that our point cloud compression approach improves performance for machine vision tasks without compromising human vision quality.

\keywords{Point Cloud Compression  \and Scalable Coding for Machine.}
\end{abstract}

\section{Introduction}
\label{sec:intro}


With the advancement of 3D data-capturing devices such as RGB-D cameras and LiDAR sensors, there is growing research interest in developing new approaches for point cloud-related machine vision tasks, including classification, segmentation, and detection. However, most of these tasks require raw point cloud data as input, leading to high bandwidth requirements for transmitting large volumes of point cloud data.

Recently, many neural compression frameworks \cite{que2021voxelcontext,chen2022point,zhou2022riddle,he2022density,fu2022octattention,hu2020improving,chen2022exploiting,chen2022lsvc,chen2021improving,chen2023neural,huang2020octsqueeze,chen2024group,han2024cra5} were proposed to meet the bandwidth/storage requirement for better point cloud data transmission/ storage. However, those point cloud compression frameworks are only designed for the human vision task without considering the performance for the machine vision tasks. While some recent image coding for machines works~\cite{9385898,9414465,song2021variable,torfason2018towards,bai2022towards,choi2022scalable,wang2021towards,liu2021semantics} have tried to optimize the network by introducing additional loss functions for the machine vision tasks, we cannot simply adopt the similar strategy (\textit{i.e.,} through adding similar loss functions) to improve the point cloud compression performance for the machine vision tasks. A possible explanation is that the mainstream point cloud compression algorithms like VoxelContext-Net\cite{que2021voxelcontext}, OctAttention~\cite{fu2022octattention}, Geometry Point Cloud Compression (G-PCC)~\cite{G-PCC}, and others~\cite{biswas2020muscle,huang2020octsqueeze,chen2022point} need to compress the pre-constructed octrees from different point clouds and the octree construction procedure is indifferentiable.  In addition, while most coding for machines methods~\cite{9385898,9414465,song2021variable,torfason2018towards,bai2022towards,liu2023icme,liu2023icmh} improve the effectiveness of various machine vision tasks, the human vision performance is still degraded. Therefore, it is desirable to design a new point cloud compression framework for machine perception, which can improve the performance for the machine vision tasks without degrading the human vision performance.

In this work, we propose a point cloud compression method that simultaneously addresses human vision and multiple machine vision tasks. Our method adheres to the scalable coding paradigm~\cite{choi2022scalable,wang2021towards,liu2021semantics}, where the entire bit-stream is used for human vision tasks, and only subsets are used for various machine vision tasks. We demonstrate the generalizability of our approach using three mainstream point cloud compression methods: VoxelContext-Net~\cite{que2021voxelcontext}, OctAttention~\cite{fu2022octattention}, and G-PCC~\cite{G-PCC}. In our method, octrees pre-constructed from different point clouds are compressed into bit-streams, effectively balancing the needs of both human and machine vision tasks.

For machine vision tasks, we introduce a bit-stream partition method that transmits only part of the bit-stream to reconstruct the initial octree depth levels, thereby saving bit-rate. Additionally, we propose an octree depth level predictor to adaptively determine the optimal octree depth level for different machine vision tasks. Consequently, for simpler tasks (\textit{e.g.}, classification) and straightforward objects/scenarios, fewer octree depth levels are used to save bit-rate. In contrast, for more complex tasks (\textit{e.g.}, segmentation) and intricate objects/scenarios, deeper octree depth levels are utilized to enhance visual recognition performance. Experimental results show that our method delivers promising outcomes across various machine vision tasks. The main contributions of this work are summarized as follows:
1). We propose combining an octree depth level predictor with a bit-stream partition method to adaptively select the optimal octree depth level and split the bit-stream for various machine vision tasks. Our method serves as a simple yet strong baseline, facilitating further research in point cloud compression for both machine and human vision.
2). Our newly proposed compression methods can be seamlessly integrated into mainstream octree-based point cloud compression frameworks, including both deep learning methods (\textit{e.g}., VoxelContext-Net and OctAttention) and handcraft methods (\textit{e.g.}, G-PCC), extending these methods for machine perception.
3). Comprehensive experiments show that our new baseline method, PCCMP-Net, delivers promising classification, detection, and segmentation results without compromising human vision performance.

\section{Related work}
\subsection{Point Cloud Compression}

In the past few years, many hand-crafted and learning-based point cloud compression methods \cite{G-PCC,wang2021lossy,chen2017neural,biswas2020muscle,huang2020octsqueeze,zhu2020view,que2021voxelcontext,fu2022octattention,chen2022point} have been proposed by transforming the point cloud data into octree representations for better compression, in which G-PCC (geometry based point cloud compression) \cite{G-PCC} proposed by the MPEG is the most popular one.

In recent years, some learning-based point cloud compression methods \cite{huang20193d,zhu2020view},\\\cite{huang2020octsqueeze,biswas2020muscle,que2021voxelcontext,wang2021lossy,wang2021multiscale,zhou2022riddle,he2022density,fu2022octattention,chen2022point,wang2022improving,yang2023neural} have achieved the state-of-the-art performance. Huang \textit{et al.} \cite{huang2020octsqueeze} and Wang \textit{et al.} \cite{wang2021lossy} followed the learned image compression framework \cite{balle2017end} to compress the voxelized point clouds. To reduce the bit-rate, Biswas \textit{et al.} \cite{biswas2020muscle} exploited the spatio-temporal relationships across multiple LiDAR sweeps by developing a new conditional entropy model. Based on \cite{wang2021lossy}, Wang \textit{et al.} \cite{wang2021multiscale} used the lossless compressed octree and the lossy compressed point feature to further improve the coding performance. To further improve point cloud compression performance, Que \textit{et al.} \cite{que2021voxelcontext} extended this framework by exploiting the context information among neighboring nodes and refining the 3D coordinate at the decoder side. Fu \textit{et al.} \cite{fu2022octattention} used masked context attention, while Chen \textit{et al.} \cite{chen2022point} used sibling context and surface priors to improve the compression performance. 

While the above-mentioned works can be readily used for downstream visual recognition tasks, these existing methods aim to compress the point cloud data only for human perception without considering the machine vision tasks. To the best of our knowledge, there is no existing point cloud compression method for both machine vision and human vision.

\subsection{Coding for Machines}

Choi \textit{et al.}~\cite{choi2022scalable} and Wang \textit{et al.}~\cite{wang2021towards} performed scalable image compression by dividing the bit-streams into different parts and transmitting parts of the bit-streams for both machine or human vision tasks, while Liu \textit{et al.}~\cite{liu2021semantics} proposed a scalable image compression method for classification. Meanwhile, Bai \textit{et al.} \cite{bai2022towards} redesigned the Vision Transformer model to improve the performance for both the image coding and the classification tasks. Yang \textit{et al.} \cite{9385898} designed the image encoder by using the edge extraction algorithm, and the reconstructed images from the decoder can achieve promising image coding for machine performance. Le \textit{et al.} \cite{9414465} directly added the additional machine vision task related losses to the compression loss functions to improve the reconstructed image quality for the machine vision tasks. Song \textit{et al.} \cite{song2021variable} compressed the source image through a corresponding quality map produced from different machine vision tasks. 
Torfason \textit{et al.} \cite{torfason2018towards} combined the image compression network with the detection network, and directly extracted the detection related information from bit-streams without using an image decoder.
Liu \textit{et al.}~\cite{liu2023icme} designed a two branches compression method namely PCHM-Net, which is the first point cloud compression method for both machine vision and human vision. However, PCHM-Net improved the machine vision performance while drop the human vision. 

In contrast to those coding for machine approachs, we propose a simple and strong baseline method to improve point cloud coding for machine perception performance. Our new method PCCMP-Net achieves promising results for machine vision tasks without sacrificing human vision performance, which cannot be achieved by the above-mentioned methods.

\begin{figure}[!t]
\begin{center}
\includegraphics[width=0.5\linewidth]{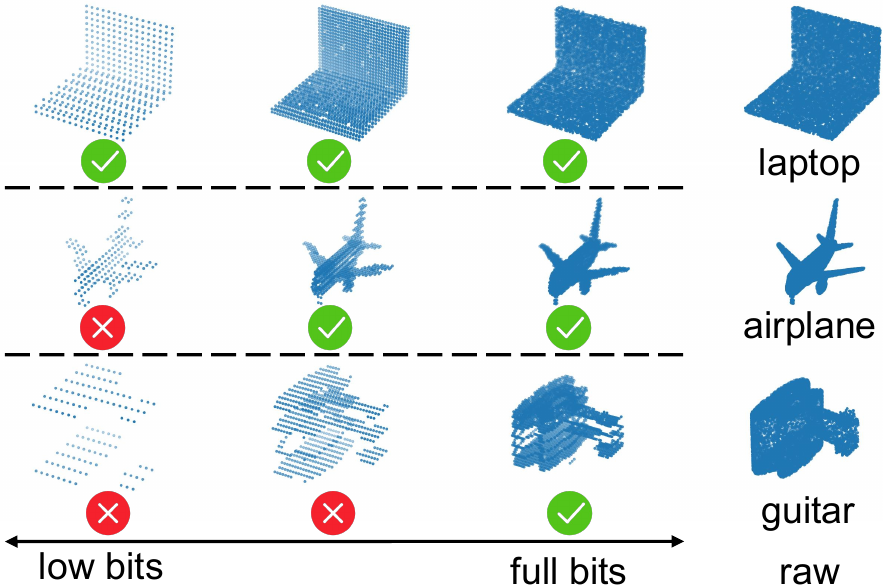}
\end{center}
\vspace{-4mm}
  \caption{The classification results of a pre-trained PointNet++\protect~\cite{qi2017pointnetplusplus} model for recognizing the point clouds reconstructed from different octree depth levels. ``raw" means the raw/original point cloud. The truly or falsely predicted results from the classification task are shown under the point clouds.}
\vspace{-3mm}
\label{fig:visual_method}
\end{figure}

\begin{figure*}[!ht]
\begin{center}
\includegraphics[width=\textwidth]{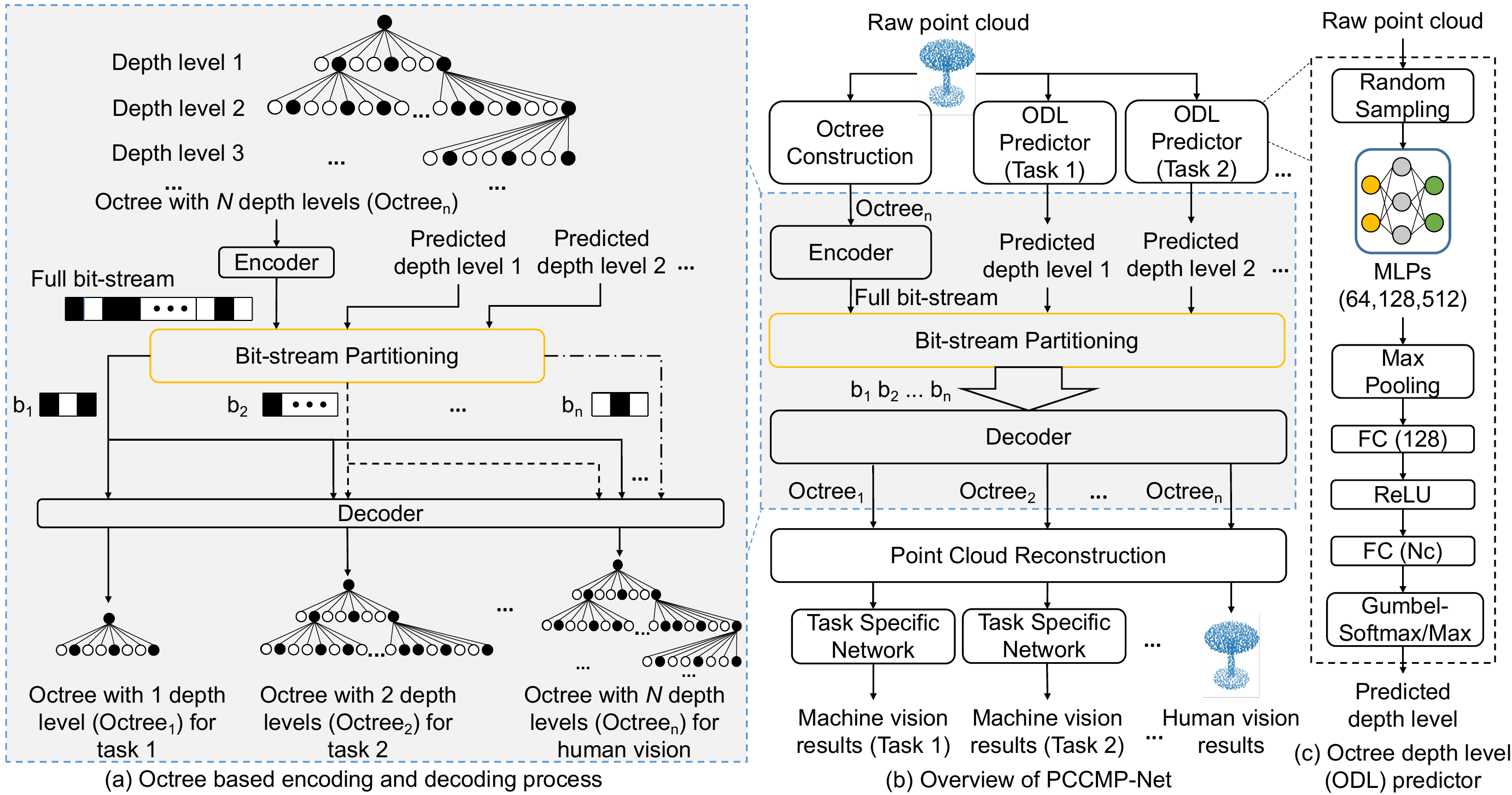}
\end{center}
\vspace{-4mm}
  \caption{(a) The octree-based encoding and decoding process. The bit-stream $\mathrm{b_1}$ is used for the first machine vision task, while the bit-stream $\mathrm{b_1} \cup \mathrm{b_2}$ will be used for the second machine vision task. And the bit-stream $\mathrm{b_1} \cup \mathrm{b_2} \cup ... \cup \mathrm{b_n}$ denotes the full bit-stream for human vision. (b) The overall network architecture of our PCCMP-Net. (c) Details of our proposed octree depth level (ODL) predictor.}
\vspace{-5mm}
\label{fig:overview}
\end{figure*}

\section{Methodology}

\subsection{Motivation}
The mainstream octree-based point cloud compression methods~\cite{huang2020octsqueeze,que2021voxelcontext,chen2022point,fu2022octattention} only use the entire bit-stream encoding information from all octree levels for the human vision task. These approaches are not optimal for machine perception. As shown in Figure~\ref{fig:visual_method},  it is unnecessary to use the entire bit-stream to recognize some objects (like laptop and airplane). Therefore, transferring the entire bit-stream for machine perception would waste the bandwidth. In order to save the bit-rate cost, it is desirable to propose a point cloud compression method for both human vision and machine vision. 

\subsection{Our New Framework}
The overall network structure of our point cloud compression method is shown in Figure~\ref{fig:overview}(b). In this section, we will first introduce the overall pipeline of our baseline method, which can well handle the human vision task and multiple machine vision tasks as shown in our experiments. And then, each module in our baseline method will be introduced.

\textbf{Overall Pipeline.} Considering that the point cloud data is commonly used for various machine vision tasks, our approach is primarily used for the machine vision tasks (\textit{e.g.,} detection of abnormal events like collision between pedestrians and vehicles) by only using subsets of bit-stream. If the human vision task must also be involved (\textit{e.g.}, when our baseline method detects abnormal events), our baseline method can additionally reconstruct a high quality point cloud with the entire bit-stream to human for further analysis. Like the scalable coding methods, when reconstructing the point clouds for the human vision task, our method will reuse the subsets of the generated bit-stream for different machine vision tasks. In this way, our point cloud compression method can improve the performance for multiple machine vision tasks without degrading the human vision performance.

\textbf{Octree Construction, Encoder, Decoder, and Point Cloud Reconstruction.} 
The octree construction module constructs the octree for each input point cloud. Octree is a tree-like data structure used to describe three-dimensional space. An octree can be constructed from any 3D point cloud by first partitioning the 3D space into 8 cubes with the same size, and then recursively partitioning each non-empty cube in the same way until the maximum depth level is reached. Each node is represented by the 3D coordinate of the cube center. Each octree is encoded as the corresponding bit-stream by using the encoder. The decoder reconstructs the octree-based on the received bit-stream. The point cloud reconstruction module then restores the point cloud from the octree. In this work, we take VoxelContext-Net~\cite{que2021voxelcontext}, OctAttention~\cite{fu2022octattention} and G-PCC~\cite{G-PCC} as the examples to introduce our method.

\textbf{Bit-stream Partitioning.} Our bit-stream partitioning module can split the full bit-stream to multi-parts bit-stream for multiple machine vision tasks and human vision task. The details are shown in section~\ref{SBP}.

\textbf{Octree Depth level Predictor.} Each of our octree depth level predictor is used to choose the optimal octree depth levels for one machine vision task, which can also guide how to split the entire bit-steam. The details of this module will be described in section \ref{ODP}.

\textbf{Task Specific Network.} To handle different point cloud based machine vision tasks, this module will use different networks. For the classification task and the segmentation task, PointNet++~\cite{qi2017pointnetplusplus} will be used in this module. For the detection task, VoteNet~\cite{qi2019votenet} and PointRCNN~\cite{Shi_2019_CVPR} is adopted in this work.

\subsection{Bit-stream Partitioning} \label{SBP}
Although we can often achieve promising performance for the human vision task by using the full bit-stream to reconstruct point clouds, it has plenty of redundant information for the machine vision tasks and thus it is less effective in terms of the bit-rate cost. Therefore, we propose a simple bit-stream partitioning method to split the bit-stream for both human and machine vision tasks.

Before introducing how to divide the bit-stream, we first introduce how to generate the point cloud bit-stream. Figure~\ref{fig:overview} (a) shows the encoding and decoding process of the octree. During the encoding process, the octree is encoded from the lower depth level to the higher depth level. Therefore, the final full bit-stream can be expressed as $\mathbf{B} = (\mathrm{b}_1, \mathrm{b}_2, ..., \mathrm{b}_n)$, where $n$ is the maximum octree depth level and $b_i$ represents the bit-stream from the $i$th depth level. At the decoder side, each octree will be reconstructed from the lower depth level to the higher depth level. Particularly, the $(i+1)$th depth level of the octree can be reconstructed with the previously reconstructed octree at the $i$ depth level and the extra bits $\mathrm{b}_{i+1}$. For example, with $\mathrm{b}_1 \cup \mathrm{b}_2$, we can reconstruct the octree with the first two depth levels, and we can readily reconstruct the octree with the first three depth levels with $\mathrm{b}_1 \cup \mathrm{b}_2 \cup \mathrm{b}_3$. 


As the example shown in Figure~\ref{fig:overview}, based on the above octree encoding and decoding process, we can split the full bit-stream $\mathbf{B}$ into multiple subsets of bit-streams $\mathrm{b}_1, \mathrm{b}_2, ..., \mathrm{b}_n$ according to the predicted octree depth levels. Particularly, the first subset of bit-stream $\mathrm{b_1}$ can be used to reconstruct the octree with the first depth level, which can be used for the machine vision task 1. By additionally using bit-stream $\mathrm{b}_2$, $\mathrm{b}_1 \cup \mathrm{b}_2$ can be used to reconstruct the octree with the first two depth levels for the machine vision task 2. We can reconstruct the \textit{N} depth levels of the octree by using the entire bit-stream $\mathrm{b}_1 \cup \mathrm{b}_2 \cup ... \cup \mathrm{b}_n$, which will be used for the human vision task. And the optimal splitting depth levels are determined by the octree depth level predictors which will be discussed below.


\subsection{Octree Depth Level Predictor} \label{ODP}

Our approach learns the octree depth level predictor to decide the optimal depth level of the octree for different machine vision tasks, which cannot only achieve the reasonable performance for various machine vision tasks but also reduce the bit-rate cost. It is worth mentioning that the encoder side (\textit{e.g.,} the LiDAR sensors) always do not have enough computing power and cannot support the complex networks. Therefore, the networks (\textit{e.g.,} PointNet++~\cite{qi2017pointnetplusplus}, VoteNet~\cite{qi2019votenet} and PointRCNN~\cite{Shi_2019_CVPR}) for handling the complex machine vision tasks are placed behind the decoder instead of the encoder side. As shown in Figure~\ref{fig:overview} (c), our octree depth level predictor is designed by using the simple three layers MLP together with two fully connected layers. 

Our octree depth level predictor can determine the optimal octree depth level for the machine vision tasks from the global feature extracted from the raw point cloud. According to different difficulty levels of the machine vision tasks, our octree depth predictor can generate $n$ probabilities $\mathbf{p} = \{p_1, p_2,..., p_n\}$ for $n$ octree depth levels, and then choose the optimal octree depth level with the highest probability. 

However, the process of choosing the octree depth level with the highest probability is non-differentiable, which makes it infeasible to train the octree depth level predictor. Therefore, we adopt the Gumbel-Softmax strategy~\cite{jang2017categorical} to address this issue. First, we generate the confidence score set $\hat{\mathbf{p}}$ from the probability set $\mathbf{p}$ with Gumbel noise as follows: 
\begin{equation}
\label{eq:gumbel noise}
\hat{p}_i = p_i + G_i, i \in \{0, 1, ..., n\}
\end{equation}
where $G_i=-\log(-\log{\epsilon})$ is the standard Gumbel noise, and $\epsilon$ is randomly sampled from a uniform distribution between 0 and 1. Therefore, we can generate the one-hot vector $\hat{\mathbf{h}} = [\hat{h}_0, \hat{h}_1, ..., \hat{h}_n]$, where $\hat{h}_i = 1$ if $i = \arg \max_{j} \hat{p}_j, j\in \{0, 1, ..., n\}$. Otherwise $\hat{h}_i = 0$. $\hat{\mathbf{h}}$ is the one-hot vector from the depth level selection process. However, the argmax operation when generating the one-hot vector makes the whole network non-differentiable. Therefore, during the backward propagation process, we apply the Gumbel-Softmax strategy and relax the one-hot vector $\hat{\mathbf{h}}$ to $\Tilde{\mathbf{h}} = [\Tilde{h}_0, \Tilde{h}_1, ..., \Tilde{h}_n]$ as follows: 
\begin{equation}
\label{eq:gumbelback}
    \Tilde{h}_i=\frac{\exp(\hat{p}_i/\tau))}{\sum_{j=0}^{n}\exp(\hat{p}_j/\tau)} , i\in \{0, 1, ...,n\}
\end{equation}
where $\tau$ is the temperature parameter and is set from 3 to 0.001 during the whole training process. Using the Gumbel-softmax strategy~\cite{jang2017categorical}, we can select the optimal depth-level of each octree for the machine vision tasks based on the argmax function during forward propagation process and approximate the gradient of the argmax function by using Eq.(\ref{eq:gumbelback}) in the back propagation process. During the inference stage, we directly select the optimal octree depth level with the maximum probability in $\mathbf{p}$.

Our approach uses multiple octree depth level predictors to simultaneously handle multiple machine vision tasks. In fact, the amount of bit-stream requested for different machine vision tasks varies. And the human vision task often requires the entire bit-stream. So we can readily use multiple octree depth level predictors for multiple machine vision tasks by splitting the entire bit-stream as multiple subsets of bit-streams, which is shown in Figure~\ref{fig:overview}. In addition, only one decoder is requested for all subsets of bit-streams instead of one specific decoder for each subset of bit-stream, which is the strategy commonly used in image coding for machines works~\cite{choi2022scalable,wang2021towards,liu2021semantics}.

\subsection{Training Strategy}

For the encoder and decoder, we directly use the pre-trained compression codecs from VoxelContext-Net~\cite{que2021voxelcontext} and OctAttention~\cite{fu2022octattention}) or hand-crafted codecs from G-PCC. For the task specific network, we directly use the pre-trained network from PointNet++~\cite{qi2017pointnetplusplus},  VoteNet~\cite{qi2019votenet}, and PointRCNN~\cite{Shi_2019_CVPR}.
When training the octree depth level predictor, we fix the parameters in the compression module and the task specific network module. The octree depth level predictors for different tasks are trained separately. To better compare the losses from the octrees with different depth levels from the machine vision tasks, we calculate the bits per point (bpp) and the task specific loss for each octree during the training process. But in the inference process, only the octree with the selected depth level will be transmitted and used for the machine vision tasks. And the loss function used for training our octree depth level predictor is shown below,
\begin{equation}
\label{eq:loss}
loss = \sum((\lambda \times \mathbf{bpp} + \mathbf{L}) \times \Tilde{\mathbf{h}})
\end{equation}

Our method selects the optimal depth level from $n$ candidate depth levels. $\mathbf{bpp} = \rm (bpp_1, bpp_2,..., bpp_n)$, and $\rm bpp$ means bits per point. $\rm bpp_i(i\in \{0, 1, ...,n\})$ represents the $\rm bpp$ for constructing the octree at the $i$th depth levels, which can be calculated  from the encoder. $\mathbf{L} = (L_1, L_2,..., L_n)$, and $L_i$ are defined as follows,
\begin{equation}
\label{eq:loss_machine}
L_i = D(f(\hat{x_i}), y_{gt}), i\in \{0, 1, ...,n\}
\end{equation}

where $f$ is the machine vision task specific network (\textit{i.e.,} PointNet++ for the classification and segmentation tasks or VoteNet and PointRCNN for the detection task). $\hat{x_i}$ is the reconstructed point cloud from the octree with $i$ depth levels. $y_{gt}$ is the ground-truth for the machine vision task. And the function $D$ can calculate the loss between $f(\hat{x_i})$ and $y_{gt}$. $\Tilde{\mathbf{h}} = [\Tilde{h}_0, \Tilde{h}_1, ..., \Tilde{h}_n]$, and $\Tilde{h}_i$ is defined in Eq.(\ref{eq:gumbelback}).
$\lambda$ in Eq.(\ref{eq:loss}) is a hyper-parameter, which is used to balance the trade off between $\mathbf{bpp}$ and $\mathbf{L}$.

\begin{figure*}[!t]
\begin{center}
\includegraphics[width=\textwidth]{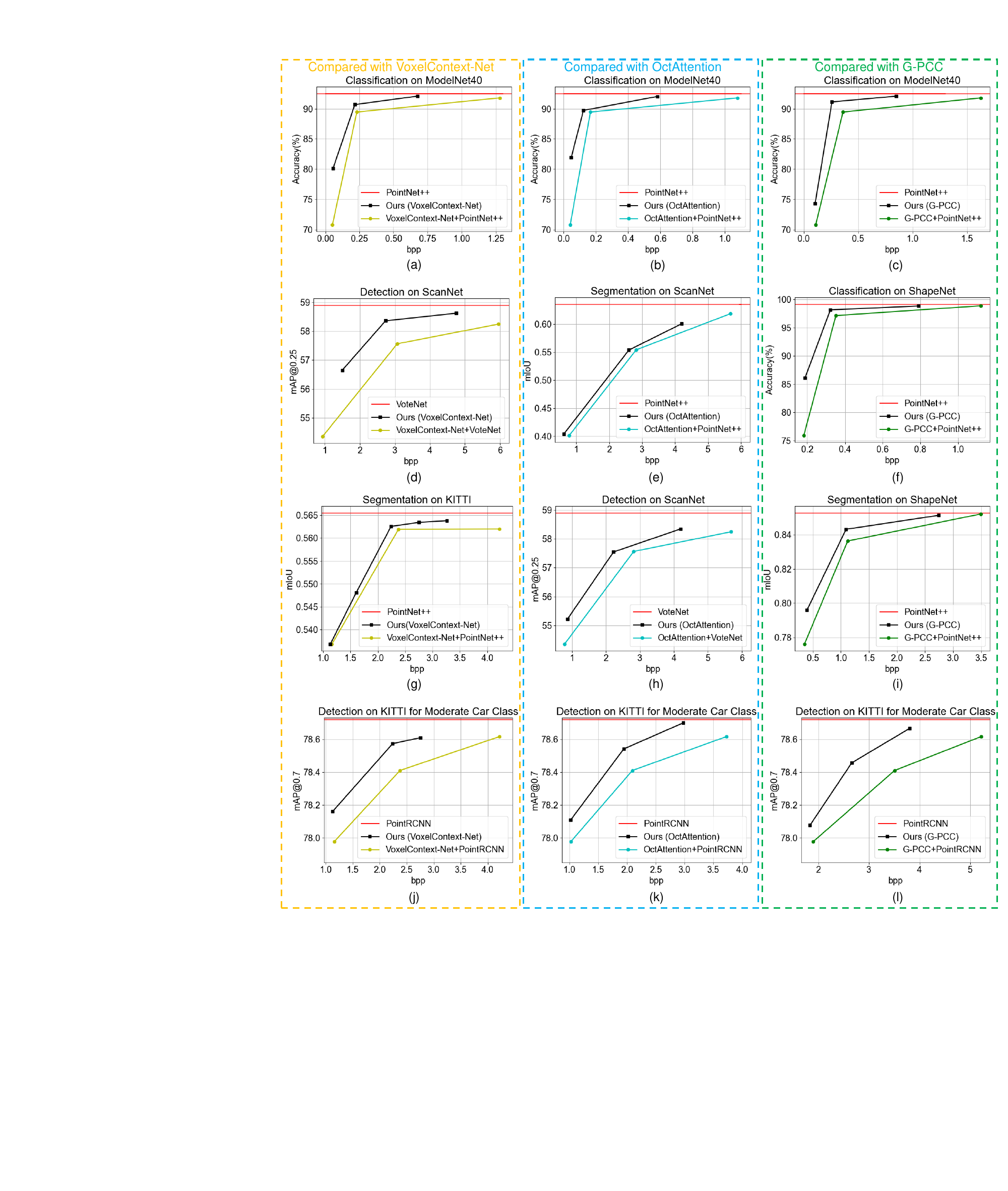}
\end{center}
\vspace{-3mm}
  \caption{Results in (a-c) are only for one single machine vision task (\textit{i.e.,} the classification task) on the ModelNet10 and ModelNet40 datasets. The multi-task results in (f), (i) are for both classification and segmentation tasks on the ShapeNet dataset. The multi-task results in (d), (e),(g), (h), (j-l) are for both segmentation and detection tasks on the ScanNet and KITTI datasets. ``Ours (VoxelContext-Net)''/``Ours (OctAttention)''/``Ours (G-PCC)'' means VoxelContext-Net, OctAttention and G-PCC are used as the encoder and the decoder in our PCCMP-Net, respectively. The results of PointNet++, VoteNet, and PointRCNN are obtained by using the raw/uncompressed point cloud as the input.}
  
\vspace{-5mm}
\label{fig:experiment_all}
\end{figure*}

\section{Experiments}




\subsection{Experimental Details}
\textbf{Baseline methods.}
Follow the ~\cite{liu2023icme}, we directly use the encoder and the decoder from VoxelContext-Net~\cite{que2021voxelcontext}, OctAttention~\cite{fu2022octattention} or G-PCC~\cite{G-PCC} as our baseline methods. We also use the same encoder and decoder for point cloud compression in our proposed framework for a fair comparison with the baseline methods.


For the baseline methods for the machine vision tasks, we directly use the reconstructed point clouds from VoxelContext-Net~\cite{que2021voxelcontext}, OctAttention~\cite{fu2022octattention} or G-PCC~\cite{G-PCC} as the input of the networks for the machine vision tasks. PointNet++~\cite{qi2017pointnetplusplus} is used for the classification task and the segmentation task.  VoteNet ~\cite{qi2019votenet} and PointRCNN~\cite{Shi_2019_CVPR} is adopted for the detection task. Note the baseline methods are referred to as the format ``compression method + task specific network'' (\textit{e.g.}, ``VoxelContext-Net+PointNet++'' means using the reconstructed point clouds from VoxelContext-Net as the input to PointNet++.)
As suggested in VoxelContext-Net~\cite{que2021voxelcontext}, we train the machine vision task networks with the raw point clouds and evaluate the classification/segmentation/detection results based on the reconstructed point clouds.

\noindent\textbf{Evaluation Metric.} We use bits per point (bpp) to denote the bit cost in the compression procedure. For the machine vision tasks, accuracy, mean intersection-over-union (mIoU) and mean average precision (mAP) are used to measure the performance for the classification, segmentation and detection tasks, respectively. For the human vision, we use the standard metric D1 PSNR (Point-to-Point PSNR) to measure the reconstructed point cloud quality.

\noindent\textbf{Implementation Details.} The whole network is implemented in Pytorch with CUDA support. At the second training stage, we set the batch size as 48. We use the Adam optimizer~\cite{kingma2015adam} with the learning rates of 1e-3 for the first 40 epochs and 1e-4 for the next 10 epochs.

\subsection{Experiment all Results}
\textbf{Results on ModelNet for Single Machine Vision Task.} As the ModelNet dataset is only designed for the classification task, our PCCMP-Net can only be used for single machine vision task on this dataset. The classification results on the ModelNet10 and ModelNet40 datasets are shown in Figure~\ref{fig:experiment_all} (a-c). When compared with the baseline method VoxelContext-Net+PointNet++, our method achieves about 10\% accuracy improvement at 0.056 bpp and saves about 45\% bpp at 91.8\% accuracy on the ModelNet40 dataset shown in Figure~\ref{fig:experiment_all} (a). In Figure~\ref{fig:experiment_all} (b) and (c), our PCCMP-Net can also achieve better performance when compared with OctAttention+PointNet++ and G-PCC+PointNet++. The experimental results demonstrate that our PCCMP-Net can improve the performance when the input point cloud is compressed for the single classification task.

\noindent\textbf{Results on ShapeNet for Multiple Machine Vision Tasks.} The experiments for multiple machine vision tasks on the ShapeNet dataset are shown in Figure~\ref{fig:experiment_all} (f) and (i), in which we use the ShapeNet dataset for both the classification task and the segmentation task. When compared with G-PCC+PointNet++ for the classification task, our PCCMP-Net achieves about 9\% accuracy improvement at the 0.2 bpp and saves about 25\% bpp at 98.9\% accuracy, as shown in Figure~\ref{fig:experiment_all} (f). For the segmentation task, our PCCMP-Net achieves 2\% mIoU improvement at lowest bpp and saves more than 20\% bpp at 0.85 mIoU, as shown in Figure~\ref{fig:experiment_all} (i). Therefore, our PCCMP-Net achieves better performance than the baseline method when simultaneously handling multiple machine vision tasks (\textit{i.e.,}the classification task and the segmentation task). 


\noindent\textbf{Results on ScanNet and KITTI for Multiple Machine Vision Tasks.} The multi-tasks experiments on the ScanNet and KITTI dataset are shown in Figure~\ref{fig:experiment_all} (d,e,g,h,j,k), in which we use the ScanNet and KITTI datasets for both segmentation task and the detection task. For the segmentation task, when compared with OctAttention+PointNet++, our PCCMP-Net saves about 10\% bpp at the 0.40 mIoU, as shown in Figure~\ref{fig:experiment_all} (e). For the detection task shown in Figure~\ref{fig:experiment_all} (h), our PCCMP-Net saves above 25\% bpp at the 0.583 mAP@0.25 when compared with OctAttention+VoteNet. In Figure~\ref{fig:experiment_all} (j), our PCCMP-Net can also achieve better performance when compared with VoxelContext-Net+PointRCNN for the KITTI dataset. These experimental results demonstrate that our PCCMP-Net can simultaneously improve the performance for multiple machine vision tasks (\textit{i.e.,}the segmentation task and the detection task).

\noindent\textbf{Human Vision Results.} 
Note that the encoder, the decoder, and the full bit-stream in our method are the same as our baseline methods VoxelContext-Net~\cite{que2021voxelcontext}, OctAttention~\cite{fu2022octattention} and G-PCC~\cite{G-PCC} in human vision.
So our PCCMP-Net achieves the same human vision performance when compared with the baseline methods. It is worth mentioning that in most coding for machines methods~\cite{liu2023icme,choi2022scalable,bai2022towards,liu2021semantics,9385898,torfason2018towards}, compression performance for human vision always drops in order to achieve better performance for the machine vision tasks. Therefore, the results indicate the advantage of our PCCMP-Net can improve the performance for the machine vision tasks without sacrificing the compression results for human vision. 




\subsection{Model Analysis}
\begin{figure}[htbp]
\begin{center}
\includegraphics[width=0.9\linewidth]{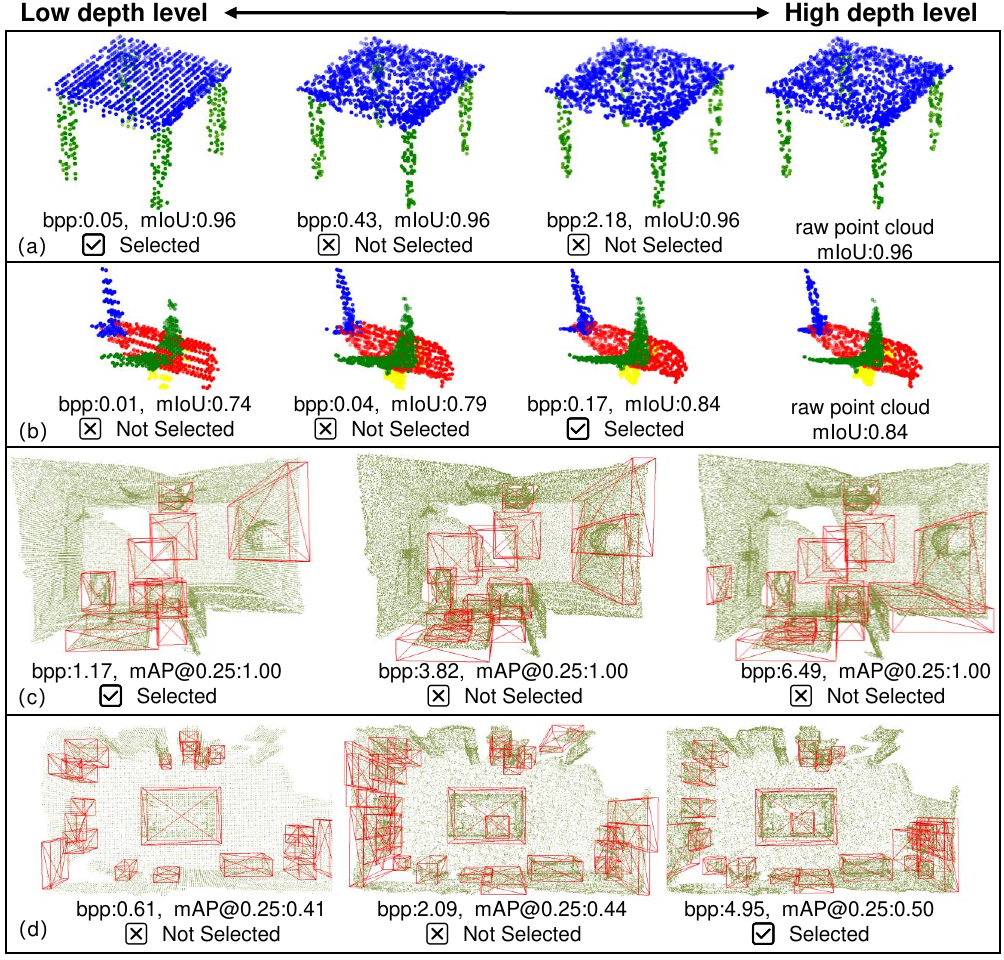}
\end{center}
\vspace{-3mm}
\caption{Different qualitative results for the segmentation task on the ShapeNet dataset (a) and (b). Different qualitative results for the detection task on the ScanNet dataset (c) and (d). }
\vspace{-5mm}
\label{fig:visualization}
\end{figure}

\begin{figure}[htbp]
\begin{center}
\includegraphics[width=0.9\linewidth]{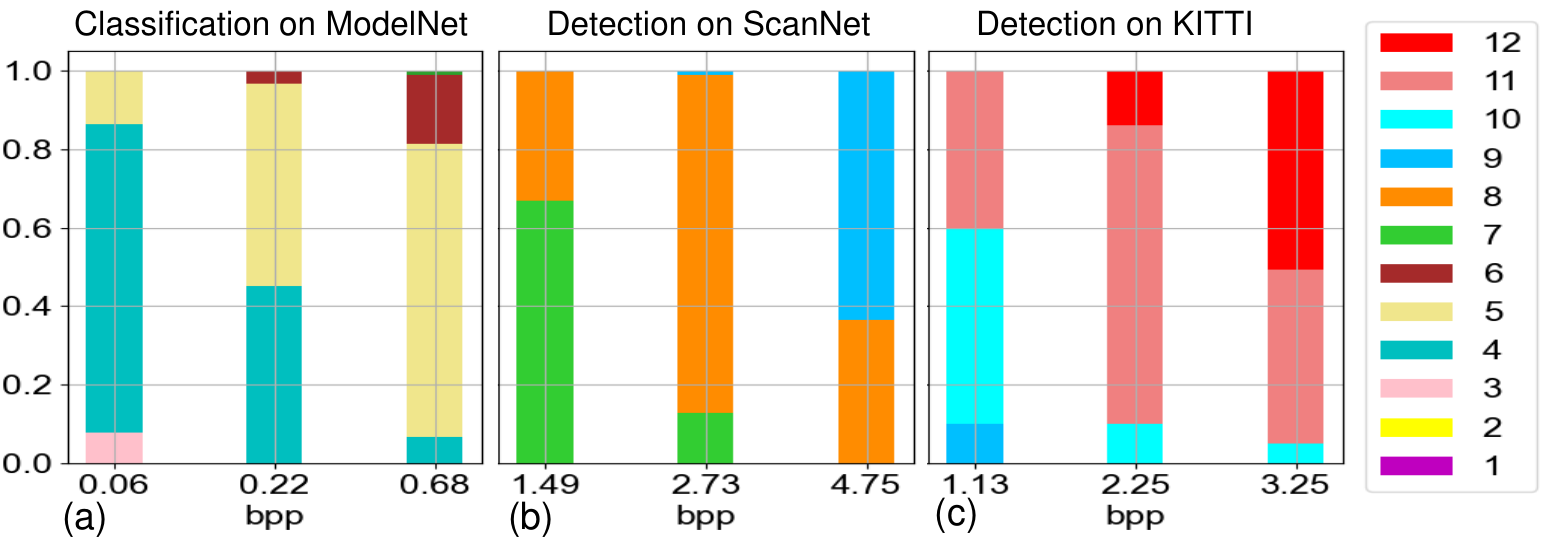}
\end{center}
\vspace{-4mm}
\caption{The selection percentage of different octree depth levels at different bpp values for the classification task on the ModelNet40 dataset (a), and the detection task on the ScanNet dataset (b) and the KITTI dataset (c). Different colors represent different depth levels.}
\vspace{-3mm}
\label{fig:diff_lambda}
\end{figure}

In order to balance the bit-rate cost and the performance of the machine vision tasks in different scenarios, our proposed octree depth level predictor can adaptively select the optimal depth levels for each point cloud. For different tasks at different bpp values, the selection percentages at different octree depth levels are shown in Figure~\ref{fig:diff_lambda}. We observe that when bpp value is smaller our method tends to select lower octree depth levels for more percentage of point clouds. With the increasing of the bpp values, our octree depth level predictor will select higher octree depth levels for more percentage of point clouds. The selection percentage in Figure~\ref{fig:diff_lambda} demonstrates that our PCCMP-Net can adaptively select the optimal octree depth level for each point cloud at different bpp values for different machine vision tasks.

The visualization results of the segmentation task and the detection task are shown in Figure~\ref{fig:visualization}. Figure~\ref{fig:visualization} (a) shows that it is easy to segment the table with only 2 regular parts, so our octree depth level predictor tends to select the lower depth level to save bits. It is harder to segment the airplane shown in Figure~\ref{fig:visualization} (b) with 4 grotesque parts, therefore our octree depth level predictor prefers the higher depth level to achieve better segmentation performance (\textit{e.g.,}mIoU). Figure~\ref{fig:visualization} (c) shows that the scene with sparse and regularly placed objects is relatively easier for the object detector, so our octree depth predictor selects the lower depth level for saving bits. On the other hand, the scene in Figure~\ref{fig:visualization} (d) with crowded and stacked objects may be challenging for the object detector, hence our octree depth level predictor tends to choose the higher depth level for better detection performance (\textit{e.g.,}mAP@0.25). It is observed that our proposed octree depth level predictor can select the optimal octree depth levels for different cases, which demonstrate the effectiveness of our proposed approach for different machine vision tasks.

\section{Conclusion}
In this work, we have proposed a new point cloud compression method for machine perception, which is referred to as PCCMP-Net. Particularly, observing that the octree-based approaches have become the main-stream point cloud compression methods, we propose to learn a set of octree depth level predictors for multiple machine vision tasks, which can be readily incorporated into both hand-crafted methods (\textit{e.g.,} G-PCC~\cite{G-PCC}) and the recent learning-based methods (\textit{e.g.,} VoxelContext-Net~\cite{que2021voxelcontext} and OctAttention~\cite{fu2022octattention}) and extend these methods towards point cloud compression for both human vision and machine vision. Comprehensive experiments on five benchmark datasets demonstrate that our newly proposed approach PCCMP-Net achieves promising results for multiple machine vision tasks (\textit{i.e.,} classification, segmentation, detection) without sacrificing the performance of the human vision task. We believe our newly proposed approach PCCMP-Net can be used as a simple and strong baseline method to facilitate the subsequent works along the emerging research area of point cloud compression for machine perception. 



\end{document}